\documentclass[]{ailab}

\usepackage{amsmath,amsfonts,bm}

\def\eqref#1{equation~\ref{#1}}

\def\1{\bm{1}}

\DeclareMathAlphabet{\mathsfit}{\encodingdefault}{\sfdefault}{m}{sl}
\SetMathAlphabet{\mathsfit}{bold}{\encodingdefault}{\sfdefault}{bx}{n}

\usepackage{booktabs}
\usepackage{amssymb}
\usepackage{pifont}
\usepackage{booktabs}
\usepackage{makecell}
\usepackage{graphicx}
\usepackage{float}
\usepackage[dvipsnames]{xcolor}

\renewcommand{\eqref}[1]{Eqn.~(\ref{#1})}

\usepackage{multirow}
\usepackage{wrapfig}
\usepackage[table]{xcolor}
\usepackage{booktabs}
\usepackage{pifont}
\usepackage{colortbl}
\usepackage{graphicx}
\usepackage{booktabs}
\usepackage{amssymb}
\usepackage{array}
\usepackage{algorithm}
\usepackage{listings}
\usepackage{xcolor}
\definecolor{codecomment}{RGB}{48,128,132}
\definecolor{icmlblue}{rgb}{0,0.08,0.45}

\lstdefinestyle{pytorch}{
    language=Python,
    basicstyle=\ttfamily\small,
    commentstyle=\color{codecomment},
    keywordstyle=\color{black},
    stringstyle=\color{black},
    showstringspaces=false,
    columns=fullflexible,
    keepspaces=true,
    breaklines=true,
    breakatwhitespace=true,
    numbers=none,
    frame=none,
    tabsize=4,
    aboveskip=5pt,
    belowskip=5pt
}
\providecommand{\cmark}{\ding{51}}
\providecommand{\xmark}{\ding{55}}

\setcitestyle{
    square,
    comma,
    numbers,
    sort&compress
}

\title{\fontsize{17pt}{19pt}\selectfont Position Forcing: Self-Conditioning 3D Generation}

\author[1,2]{Ziheng Ouyang}
\author[2,3]{Zeqiang Lai\textsuperscript{\(\dagger\)}}
\author[2,4,5]{Jiarui Chen}
\author[2,3]{Jiangshan Wang}
\author[1]{Yuhao Wan}
\author[1]{\\ Jingbo Gong}
\author[3]{Xiangyu Yue}
\author[6]{Hengshuang Zhao}
\author[1]{Qibin Hou\textsuperscript{\(\ddagger\)}}
\author[2]{Chunchao Guo\textsuperscript{\(\ddagger\)}}

\affiliation[1]{VCIP, Nankai University\quad}
\affiliation[2]{Tencent Hunyuan\quad}
\affiliation[3]{MMLab, CUHK\\}
\affiliation[4]{Fudan University\quad}
\affiliation[5]{Shanghai Innovation Institute\quad}
\affiliation[6]{HKU}

\contribution{\textsuperscript{\(\dagger\)} Project lead.}
\contribution{\textsuperscript{\(\ddagger\)} Corresponding author.}

\resource{\faGithub\quad\href{https://ouyangziheng.github.io/Position-Forcing-Page/}{Project Website}}

\email{zihengouyang666@gmail.com, andrewhoux@gmail.com,
chunchaoguo@tencent.com}

\abstract{
Recent single-stage 3D generative models commonly adopt VecSet representations, encoding 3D shapes as unordered sets of latent tokens.
However, compared with two-stage methods that provide explicit positional guidance, these models must implicitly infer token positions throughout denoising, limiting their generation quality.
We observe that, despite the absence of explicit positional conditioning, VecSet tokens retain recoverable spatial correspondences.
Building on this observation, we propose Position Forcing, a position-based self-conditioning framework. During denoising, Position Forcing recovers token positions from the current clean latent estimate, quantizes them at progressively finer resolutions according to the denoising stage, and feeds the resulting positional encodings back into the diffusion Transformer.
This progressively refined positional feedback provides spatial guidance at a granularity appropriate to each denoising stage, guiding shape generation along a coarse-to-fine trajectory and substantially improving generation quality without a separate position generation stage.
Experiments demonstrate that Position Forcing achieves strong performance among single-stage 3D generative methods and outperforms several competitive multi-stage approaches.
\par
\vspace{0.2em}
{\centering

\par}
\par

}

\begin{document}

\thispagestyle{firstheader}

\maketitle


\section{Introduction}

3D content is a fundamental component of the digital world, supporting a wide range of applications in gaming, filmmaking, and virtual reality~\citep{xu2024sketch2scene,wang2026dreamstyle3d,gong2026direct}.
However, creating high-quality 3D assets typically requires specialized modeling expertise and substantial manual effort, making it difficult to meet the growing demand for 3D content.
Recent advances in generative models have enabled the automatic synthesis of high-quality 3D assets conditioned on images or text, offering new opportunities to make 3D creation more accessible and efficient.
In particular, diffusion- and flow-based generative models~\citep{ho2020denoisingdiffusionprobabilisticmodels,rombach2022highresolutionimagesynthesislatent,lipman2023flowmatchinggenerativemodeling,peebles2023scalablediffusionmodelstransformers,ma2024sitexploringflowdiffusionbased,wang2026elasticdiffusiontransformer,zhou2025onevaejointdiscretecontinuous,ouyang2026consistency} have been widely adopted for 3D geometry generation, substantially improving generation quality and practical usability~\citep{zhao2023michelangelo,hunyuan3d2.0,li2025triposg,wang2026does}.

To efficiently model complex 3D geometry, recent 3D generative methods widely adopt VecSet-style representations, which compress shapes into a compact set of latent tokens and enable generative modeling in an efficient latent space~\citep{zhang20233dshape2vecset,zhao2023michelangelo,craftsman3d,hunyuan3d2.0}.
During generation, however, the model starts from a set of unordered noise tokens.
Without explicit positional guidance, the model must establish both the spatial organization of the tokens and their corresponding geometric representations within a single denoising process, making generative modeling more challenging.

To alleviate this difficulty, recent methods adopt multi-stage generation paradigms and achieve substantial improvements in generation quality~\citep{trellis,lattice,trellis2}.
These methods typically first generate a coarse spatial structure to explicitly determine where geometric content should reside, and then use this structure to guide fine-grained geometry generation.
This explicit spatial guidance facilitates subsequent geometry
modeling but relies on a separate structure-generation model
and an additional sampling stage.
We therefore seek to obtain explicit spatial guidance directly within a single-stage generation process, without introducing an additional structure-generation stage.

\begin{figure}
    \centering
    \includegraphics[width=\linewidth]{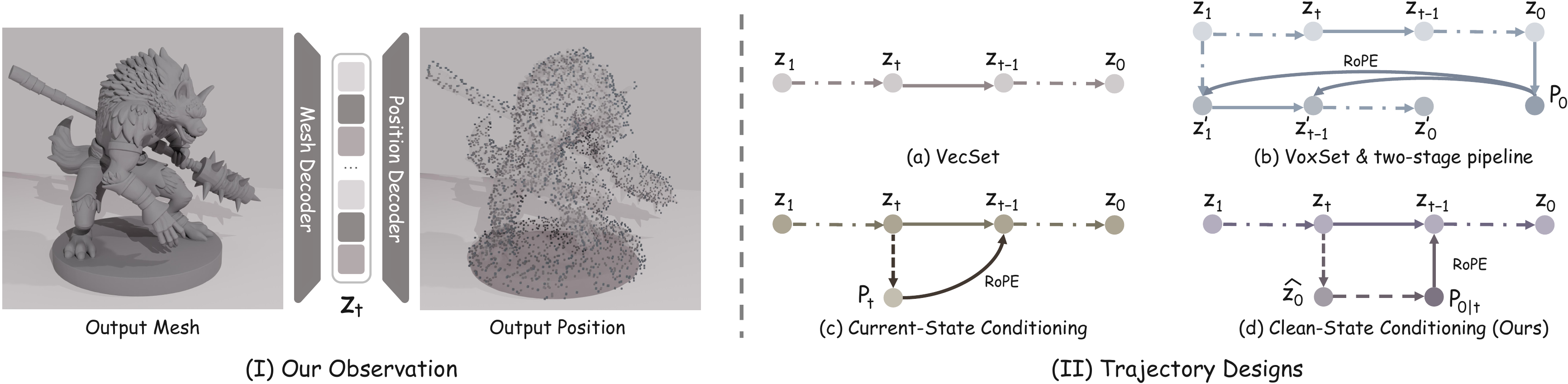}
    \vspace{-12pt}
    \caption{
        \textbf{Our observation and positional conditioning designs.}
        \textbf{(I)} Latents can be decoded into shape geometry and token positions.
        \textbf{(II)} (a) VecSet uses no explicit positional conditioning; (b) VoxSet and two-stage pipelines use positions established beforehand.
        (c) Our naive design recovers positions from the current noisy state.
        (d) \emph{Position Forcing} recovers positions from the predicted clean state and applies progressive quantization to provide coarse-to-fine spatial guidance.
    }
    \label{fig:overview}
    \vspace{-12pt}
\end{figure}

To this end, we further investigate the spatial properties of VecSet latent representations.
In a VecSet VAE with point queries, each latent token is constructed from a corresponding spatial query and is therefore associated with a query position.
We observe that this positional correspondence is retained in the encoded latent representation.
As shown in Fig.~\ref{fig:overview}(I), the corresponding query positions can be recovered from the latent tokens alone, indicating that the unordered VecSet representation retains decodable spatial information.

Since this positional information can be recovered from latent tokens, a natural idea is to feed the recovered positions back into the model as explicit spatial conditioning during generation.
To this end, we first consider a naive design that predicts token positions from the current noisy latent during both training and inference and constructs positional encodings at full spatial resolution, as illustrated in Fig.~\ref{fig:overview}(II)(c).
However, we find that this direct approach often produces shapes with incorrect geometry.
At high noise levels, position predictions are unreliable, and full-resolution positional encodings can make the model rely on incorrect fine-grained spatial relationships.

Building on these observations, we propose \textbf{Position Forcing}, a position-based self-conditioning framework that provides spatial guidance through progressively refined positional feedback.
We first introduce a progressive position quantization strategy governed by the noise level, matching the granularity of spatial conditioning to the current generation step.
At high noise levels, we quantize positions onto a coarse spatial grid to preserve global layout information while reducing sensitivity to positional errors.
As the noise level decreases, we progressively increase the quantization resolution, refining positional guidance from global layout to local spatial relationships.

To further reduce the impact of noise on position recovery and provide more accurate positional guidance, we consider recovering token positions from the predicted clean latent representation.
Specifically, during inference, we use the model-predicted velocity field to compute the predicted clean latent, recover token positions from it, and progressively quantize them, as illustrated in Fig.~\ref{fig:overview}(II)(d).
These positional conditions are fed back into the diffusion Transformer through RoPE, providing coarse-to-fine spatial guidance for subsequent denoising.
To further improve generation quality, we train the model with accurate query positions using the same quantization.
Together, these designs enable Position Forcing to continually update spatial guidance within a single denoising trajectory, guiding generation from global layout to fine-grained local geometry without a separate position generation stage.

Our main contributions are summarized as follows:

\begin{itemize}
    \item We reveal that VecSet latent tokens retain decodable spatial information, allowing their corresponding query positions to be recovered with high accuracy.

    \item We propose \textbf{Position Forcing}, a self-conditioning framework that combines position quantization with position recovery from predicted clean latents to provide increasingly precise spatial guidance.

    \item Our method guides generation along a coarse-to-fine spatial trajectory, enabling high-quality single-stage 3D generation without a separate position generation stage.
\end{itemize}

\section{Related Work}

\textbf{Shape Generation.}
3D shape generation spans GAN/VAE models, 2D diffusion priors,
feed-forward reconstruction, and optimization-based methods.~\citep{hong2024lrm,yang2024hunyuan3d,tang2024lgm,chen2026megs2memoryefficientgaussiansplatting},
and autoregressive mesh modeling~\citep{siddiqui2024meshgpt,weng2025scaling,hao2024meshtron}.
Methods generating directly in 3D representation spaces
broadly follow two lines.
The first encodes geometry into compact VecSet latent tokens,
as in 3DShape2VecSet~\citep{zhang20233dshape2vecset},
Michelangelo~\citep{zhao2023michelangelo},
CLAY~\citep{zhang2024clay},
Hunyuan3D 2.0~\citep{hunyuan3d2.0},
TripoSG~\citep{li2025triposg}, and Step1X-3D~\citep{li2025step1x}.
These methods perform generation in a compressed latent space and recover geometric details through a learned decoder.
LATTICE~\citep{lattice} adds explicit spatial structure through
voxel anchoring and layout-first generation.
The second models local geometry on spatially structured sparse
grids, including XCube~\citep{ren2024xcube},
TRELLIS~\citep{trellis}, Hi3DGen~\citep{hi3dgen},
SparseFlex~\citep{he2025sparseflex}, Sparc3D~\citep{li2026sparc3d},
Direct3D-S2~\citep{direct3d}, and TRELLIS 2~\citep{trellis2}.

\textbf{Generation Ordering and Self-Conditioning.}
Generation ordering structures information emergence through
noise schedules or generation hierarchies.
Diffusion Forcing~\citep{chen2024diffusion},
Latent Forcing~\citep{baade2026latent}, and
Self-Flow~\citep{chefer2026self} explore differentiated noise
or information levels, while VAR~\citep{tian2024visual},
FlowAR~\citep{ren2024flowar}, NVG~\citep{wang2026next}, and
Trajectory Forcing~\citep{kocabas2026trajectory} organize
generation through spatial scales, token granularity,
or structured intermediate states.
Such designs allow earlier structural information to serve
as context for subsequent detail generation.
Complementarily, Analog Bits~\citep{chen2022analog} and
RIN~\citep{jabri2022scalable} perform self-conditioning by
reusing clean-sample predictions and internal latent states,
respectively.
These mechanisms allow subsequent denoising steps to build on
previously computed information and progressively refine the output.

\begin{figure}
    \centering
    \includegraphics[width=1\linewidth]
    {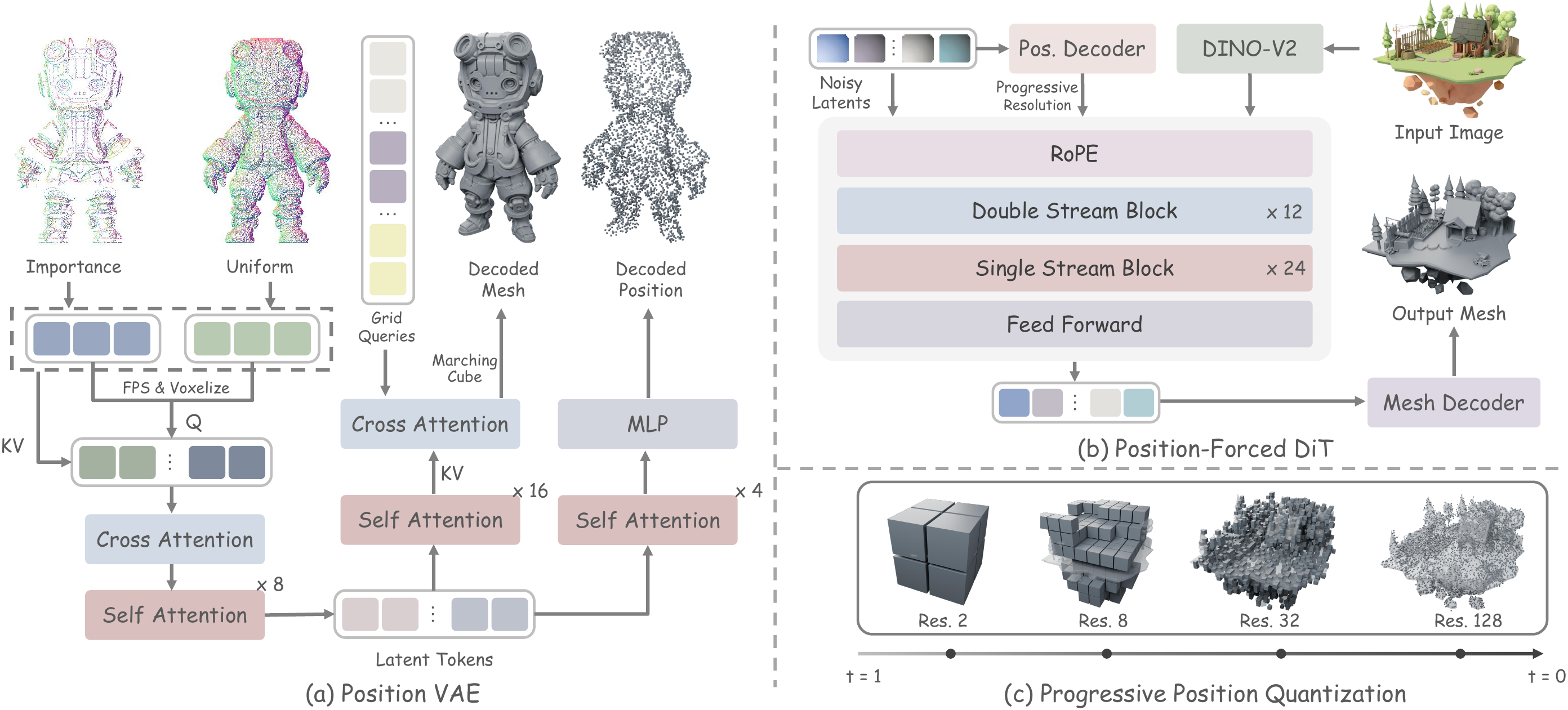}
    \vspace{-14pt}
\caption{
    \textbf{Method overview.}
    \textbf{(a)} Position VAE learns latent tokens that preserve geometry--position correspondences, enabling both shape geometry and token-associated spatial positions to be decoded.
    \textbf{(b)} During denoising, Position-Forced DiT recovers token positions from the predicted clean latents and injects them as positional guidance into subsequent denoising steps through RoPE.
    \textbf{(c)} The recovered positions are progressively quantized from coarse to fine as denoising proceeds.
}
    \vspace{-7pt}
    \label{fig:method}
\end{figure}


\section{Method}
\label{sec:method}

As illustrated in Fig.~\ref{fig:method}, our framework comprises two main components: Position VAE and Position-Forced DiT.
Position VAE augments the geometry decoder with a position decoder that recovers the spatial positions associated with latent tokens.
During denoising, Position-Forced DiT uses this position decoder to recover positions from the current clean latent estimate. These positions are quantized at progressively finer resolutions as the noise level decreases and fed back into the DiT through RoPE, providing coarse-to-fine spatial guidance for single-stage 3D generation.

\subsection{Preliminaries}
\label{sec:preliminaries}
\paragraph{VoxSet VAE.}
Conventional point-query VecSet VAEs encode an input geometry
into a compact set of latent tokens using spatial queries sampled
from the object surface.
LATTICE~\citep{lattice} then introduces VoxSet by replacing point
queries with voxel queries located at the centers of active voxels,
thereby anchoring latent tokens to a coarse spatial grid.
To accommodate voxel queries at different spatial resolutions,
LATTICE perturbs the surface queries selected by farthest point
sampling (FPS) before encoding:
\begin{equation}
\mathbf{p}_i
=
\mathbf{p}_i^{\mathrm{FPS}}+\boldsymbol{\delta}_i,
\qquad
\boldsymbol{\delta}_i
\sim
\mathcal{U}
\left[
-\frac{1}{2R},
\frac{1}{2R}
\right]^3,
\label{eq:query_jitter}
\end{equation}
where $R$ denotes the spatial resolution and $\mathbf{p}_i$
is the query position supplied to the encoder.
The encoder produces latent tokens associated with these queries
through cross-attention with the input point cloud followed by
self-attention.
\paragraph{Rectified Flow.}
We adopt rectified flow in the VAE latent space, learning a velocity field that progressively transforms Gaussian noise into clean geometry latents.
Given a clean latent $\mathbf{Z}_0$ and Gaussian noise
$\boldsymbol{\epsilon}$, the noisy state used during training
is constructed by linear interpolation:
\begin{equation}
    \mathbf{Z}_t
    =
    (1-t)\mathbf{Z}_0+t\boldsymbol{\epsilon},
    \qquad
    \boldsymbol{\epsilon}\sim\mathcal{N}(\mathbf{0},\mathbf{I}),
    \quad t\in[0,1].
    \label{eq:flow_path}
\end{equation}
Here, $t=0$ corresponds to clean latents and $t=1$ to pure noise,
with target velocity $\boldsymbol{\epsilon}-\mathbf{Z}_0$.
Given conditioning $\mathbf{c}$, a Transformer $v_\theta$ takes
the current noisy state and timestep as input and learns to
predict this velocity through the flow-matching objective:
\begin{equation}
    \mathcal{L}_{\mathrm{FM}}
    =
    \mathbb{E}_{\mathbf{Z}_0,\boldsymbol{\epsilon},t,\mathbf{c}}
    \left[
        \left\|
            v_\theta(\mathbf{Z}_t,t,\mathbf{c})
            -
            (\boldsymbol{\epsilon}-\mathbf{Z}_0)
        \right\|_F^2
    \right].
    \label{eq:flow_matching}
\end{equation}
At inference time, generation starts from Gaussian noise and
integrates the predicted velocity field from $t=1$ to $t=0$
to obtain the generated geometry latent.

\subsection{Position VAE}
\label{sec:position_vae}
\begin{wraptable}{r}{0.42\linewidth}
\vspace{-14pt}
\centering

\caption{
Effect of joint training on position recovery.
Accuracy (\%) measures the fraction of predicted positions
that fall into the same cells as their corresponding query
positions on a $128^3$ grid.
}
\label{tab:position_recovery}

\setlength{\tabcolsep}{4pt}
\resizebox{\linewidth}{!}{
\begin{tabular}{lcccc}
\toprule
& \multicolumn{4}{c}{\textbf{Noise Level (\%)}} \\
\cmidrule(lr){2-5}
\textbf{Training Strategy}
& \textbf{0}
& \textbf{10}
& \textbf{20}
& \textbf{30} \\
\midrule

Frozen VAE
& 67.4 & 51.3 & 23.0 & 6.7 \\

Joint Finetuning
& \textbf{99.9} & \textbf{99.7}
& \textbf{92.5} & \textbf{67.3} \\

\bottomrule
\end{tabular}
}
\vspace{-18pt}
\end{wraptable}

We first reproduce and pretrain the VoxSet VAE described in LATTICE~\citep{lattice}, then augment it with a Transformer-based position decoder to recover the spatial positions associated with latent tokens, yielding Position VAE.
We adopt a two-stage training strategy: first training the position decoder with the VAE frozen, then jointly finetuning the VAE and the position decoder.
\paragraph{Stage I: Position Learning on Frozen Latents.}
Given an input geometry $\mathcal{X}$, we denote the spatial queries
supplied to the encoder after query jitter by
$\mathcal{P}=\{\mathbf{p}_i\}_{i=1}^{N}$.
The pretrained VAE encoder produces a set of latent tokens:
\begin{equation}
    \mathbf{Z}
    =
    E(\mathcal{X},\mathcal{P})
    =
    \{\mathbf{z}_i\}_{i=1}^{N}.
\end{equation}
Each latent token $\mathbf{z}_i$ corresponds to the spatial query
$\mathbf{p}_i$ used to construct it, allowing these jittered query
coordinates to serve directly as token-wise positional supervision
without additional matching.
In this stage, we freeze all parameters of the VAE encoder and
geometry decoder and train only a Transformer-based position
decoder $D_{\mathrm{pos}}$.
The position decoder takes the complete latent token set, models
inter-token relationships through self-attention, and predicts
the corresponding 3D position for each token using an MLP:
\begin{equation}
    \widehat{\mathbf{P}}
    =
    D_{\mathrm{pos}}(\mathbf{Z})
    =
    \{\widehat{\mathbf{p}}_i\}_{i=1}^{N},
    \qquad
    \widehat{\mathbf{p}}_i \in \mathbb{R}^{3}.
\end{equation}
We supervise position prediction using a token-wise L1 loss:
\begin{equation}
    \mathcal{L}_{\mathrm{pos}}
    =
    \frac{1}{3N}
    \sum_{i=1}^{N}
    \left\|
        \widehat{\mathbf{p}}_i-\mathbf{p}_i
    \right\|_1.
    \label{eq:position_loss}
\end{equation}
This stage allows the position decoder to learn to extract spatial
information already present in the pretrained latent representations
while leaving the original geometry latent space unchanged.
However, as shown in Table~\ref{tab:position_recovery}, training
the position decoder on frozen latent representations alone yields
limited position prediction accuracy, which further degrades
under noise perturbations.
\paragraph{Stage II: Joint Position-Aware Training.}
In the second stage, we therefore unfreeze the VAE and jointly finetune the encoder, geometry decoder, and position decoder so that the latent representations support both geometry reconstruction and position recovery.
We retain the original VAE training objective, comprising a TSDF-based geometry reconstruction loss and KL regularization:
\begin{equation}
    \mathcal{L}_{\mathrm{VAE}}
    =
    \mathcal{L}_{\mathrm{rec}}
    +
    \lambda_{\mathrm{KL}}\mathcal{L}_{\mathrm{KL}},
    \label{eq:vae_loss}
\end{equation}
where $\mathcal{L}_{\mathrm{rec}}$ follows the TSDF mean squared error reconstruction objective of the original VAE.
We incorporate the positional supervision from Stage I to obtain the joint training objective:
\begin{equation}
    \mathcal{L}_{\mathrm{joint}}
    =
    \mathcal{L}_{\mathrm{VAE}}
    +
    \lambda_{\mathrm{pos}}\mathcal{L}_{\mathrm{pos}}.
    \label{eq:joint_position_vae}
\end{equation}
Joint training makes the corresponding token positions
easier to recover from the latent representations.
As shown in Table~\ref{tab:position_recovery}, despite being trained only on clean latent tokens, the position decoder maintains high position recovery accuracy under noise perturbations after joint finetuning.
This robustness supports the use of the position decoder to recover positional conditions from imperfect clean latent estimates in the subsequent Position-Forced DiT.
Joint finetuning also improves downstream generation quality, as demonstrated in our ablation study (Sec.~\ref{sec:ablation}).

\subsection{Position-Forced DiT}
\label{sec:position_dit}

As illustrated in Fig.~\ref{fig:method}(b), Position-Forced DiT provides explicit spatial guidance for geometry generation through progressively refined positional conditions.
Following the image-conditioned rectified-flow Transformer architecture of Hunyuan3D-2.1~\citep{hunyuan3d2.1} and LATTICE~\citep{lattice}, we extract image features using DINOv2, concatenate them with the geometry tokens, and jointly process the resulting sequence using a FLUX-style~\citep{flux} Transformer.
Positional conditions are injected into Transformer attention
blocks through 3D RoPE to encode spatial relationships among
geometry tokens.
\paragraph{Progressive Position Quantization.}
A straightforward approach is to predict token positions from the
current noisy latent during both training and inference and construct
positional conditions at full spatial resolution.
As illustrated in Fig.~\ref{fig:overview}(II)(c), this
\emph{Current-State Conditioning} recovers positions directly
from the current latent:
\begin{equation}
    P_t
    =
    D_{\mathrm{pos}}(\mathbf{Z}_t).
    \label{eq:current_position_recovery}
\end{equation}
The recovered positions are then used to construct PE and fed
back into the generative model.
However, we find that this direct approach often produces shapes
with incorrect geometry, as full-resolution PE conditions the
model on inaccurate spatial positions at high noise levels.
We therefore introduce progressive position quantization to adjust
the granularity of spatial conditioning according to the current
noise level.
Specifically, we linearly interpolate the resolution exponent
and round it down, then clamp the spatial resolution to a
specified range:
\begin{equation}
    m(t)
    =
    \left\lfloor
        t m_{\min}+(1-t)m_{\max}
    \right\rfloor,
    \qquad
    R(t)
    =
    \operatorname{clip}
    \left(2^{m(t)},R_{\min},R_{\max}\right).
    \label{eq:resolution_schedule}
\end{equation}
Here, $m_{\min}$ and $m_{\max}$ are the endpoints of the exponent
interpolation, while $R_{\min}$ and $R_{\max}$ specify the
coarsest and finest spatial resolutions, respectively.
As illustrated in Fig.~\ref{fig:method}(b), coarse quantization
at high noise levels retains global spatial layout while reducing
sensitivity to positional errors. As noise decreases, progressively
finer grids provide more precise local spatial relationships.

\paragraph{Clean-State Position Recovery and Conditioning.}
To further reduce the impact of noise on position recovery and
provide more accurate positional guidance, we consider recovering
token positions from the predicted clean latent representation.
Following Eq.~\ref{eq:flow_path}, we compute the predicted clean
latent using the model-predicted velocity $\widehat{\mathbf{V}}_t$:
\begin{equation}
    \widehat{\mathbf{Z}}_0
    =
    \mathbf{Z}_t-t\widehat{\mathbf{V}}_t.
    \label{eq:clean_latent_estimate}
\end{equation}
Here, $\widehat{\mathbf{Z}}_0$ denotes the clean latent prediction
at the current timestep.
We then recover token positions using the frozen position decoder:
\begin{equation}
    P_{0\mid t}
    =
    D_{\mathrm{pos}}
    \left(
        \widehat{\mathbf{Z}}_0
    \right).
    \label{eq:clean_position_recovery}
\end{equation}
As illustrated in Fig.~\ref{fig:overview}(II)(d), this
\emph{Clean-State Conditioning} uses positions recovered from the
predicted clean state, together with the progressive quantization
described above, to provide spatial guidance for subsequent denoising.
\paragraph{Training and Inference.}
During inference, each step uses positions derived
from the preceding step, whereas standard flow-matching
training samples individual timesteps without the corresponding
generation history.
Exactly reproducing this feedback during training requires
unrolling the preceding sampling trajectory, introducing
additional computational overhead.
We therefore retain single-timestep training and use the
ground-truth query positions $\mathcal{P}$ associated with
the VAE encoder to provide an accurate spatial reference.
These positions are progressively quantized using the same
resolution schedule $R(t)$.
Here, $Q_R$ partitions the normalized space $[-1,1]^3$ into
$R^3$ cells and maps each position to the representative
coordinates of its cell.
To maintain a consistent coordinate scale across quantization
levels, we express these representative coordinates in the
coordinate system of the maximum resolution $R_{\max}$,
with this mapping included in the definition of $Q_R$.
During training, we add random perturbations $\boldsymbol{\delta}$
to the quantized and mapped coordinates to improve robustness
to position prediction errors at inference and coordinate
jumps caused by changes in quantization resolution.
When $R=1$, we disable the perturbations so that all tokens
share the same positional encoding.

During DiT training, we freeze Position VAE and optimize
the Transformer with the following flow-matching objective:
\begin{equation}
    \mathcal{L}_{\mathrm{FM}}
    =
    \mathbb{E}_{(\mathbf{Z}_0,\mathcal{P},I),
    \boldsymbol{\epsilon},t,\boldsymbol{\delta}}
    \left[
        \left\|
            v_{\theta}
            \left(
                \mathbf{Z}_t,t,\mathbf{c}_I,
                \operatorname{PE}
                \left(
                    Q_{R(t)}(\mathcal{P})
                    +\boldsymbol{\delta}
                \right)
            \right)
            -
            (\boldsymbol{\epsilon}-\mathbf{Z}_0)
        \right\|_F^2
    \right].
    \label{eq:position_flow_matching}
\end{equation}
Here, $\mathbf{c}_I$ denotes the DINOv2 features of the input
image $I$, $\operatorname{PE}$ denotes positional conditioning
implemented through 3D RoPE, and $\boldsymbol{\delta}$ is
applied only during training.
At inference time, we construct spatial conditions from
positions recovered from the clean latent predicted at
the preceding step, without position perturbations.
At the initial pure-noise step, we set $R=1$ so that all tokens
share the same positional encoding, allowing sampling to begin
without a preceding position prediction.
Below, $t$ and $t-1$ are used as shorthand for consecutive
sampling times.
At step $t$, we use the predicted velocity to compute the
clean latent prediction and recover positions $P_{0\mid t}$.
These positions are then quantized at the next step's
resolution to condition its velocity prediction:
\begin{equation}
    \widehat{\mathbf{V}}_{t-1}
    =
    v_{\theta}
    \left(
        \mathbf{Z}_{t-1},t-1,\mathbf{c}_I,
        \operatorname{PE}
        \left(
            Q_{R(t-1)}(P_{0\mid t})
        \right)
    \right).
    \label{eq:position_forced_velocity}
\end{equation}
We iteratively update the positional conditions during sampling
and decode the resulting latent tokens into a mesh using
the geometry decoder of Position VAE.
We further evaluate these positional conditioning designs
in our ablation study (Sec.~\ref{sec:ablation}).


\section{Experiments}

\subsection{Reconstruction}
\begin{wraptable}{r}{0.43\linewidth}
\vspace{-25pt}
\centering
\caption{Quantitative comparisons of geometry reconstruction.}
\label{tab:reconstruction}
\vspace{-8pt}
\resizebox{\linewidth}{!}{
\begin{tabular}{l|c|cc}
\toprule
\textbf{Method} &
\textbf{Latent Size} &
\textbf{CD$\downarrow$} &
\textbf{F1$\uparrow$} \\
\midrule

TripoSG
& $64 \times 4096$ & 17.96 & 92.33 \\
{\footnotesize \citep{li2025triposg}}
& $64 \times 8192$ & 16.61 & 94.19 \\
\midrule

Hunyuan3D-2.1
& $64 \times 4096$ & 9.08 & 89.43 \\
{\footnotesize \citep{hunyuan3d2.1}}
& $64 \times 8192$ & 8.28 & 90.99 \\
&
$64 \times 20480$ & 7.62 & 92.06 \\
\midrule

\textbf{Position Forcing}
& $64 \times 4096$ & 8.28 & 90.74 \\
\textbf{(Ours)}
& $64 \times 8192$ & 6.37 & 93.83 \\
&
$64 \times 20480$
& \textbf{5.39} & \textbf{95.38} \\

\bottomrule
\end{tabular}
}

\vspace{-8pt}
\end{wraptable}

We evaluate the reconstruction capability of our VAE on a held-out set of meshes excluded from training, comparing against TripoSG~\citep{li2025triposg} and Hunyuan3D-2.1~\citep{hunyuan3d2.1} at different latent sizes.
We report Chamfer Distance (CD) and F1 score, where lower CD and higher F1 indicate better reconstruction quality.
As shown in Table~\ref{tab:reconstruction}, increasing the latent size consistently improves reconstruction performance, with our largest latent configuration achieving the best results.

\subsection{Generation}
\begin{table}[t]
\centering
\caption{
Comparison with existing 3D generation methods.
Best results are highlighted in bold.
}
\label{tab:generation}

\small
\setlength{\tabcolsep}{6.5pt}
\renewcommand{\arraystretch}{1.12}

\resizebox{\columnwidth}{!}{%
\begin{tabular}{l|c|cccc}
\toprule
\textbf{Model}
& \textbf{Single Stage}
& \textbf{ULIP-T}$\uparrow$
& \textbf{ULIP-I}$\uparrow$
& \textbf{Uni3D-T}$\uparrow$
& \textbf{Uni3D-I}$\uparrow$ \\
\midrule

TRELLIS~\citep{trellis}
& \xmark
& 0.076 & 0.126 & 0.249 & 0.311 \\

TRELLIS 2~\citep{trellis2}
& \xmark
& \textbf{0.077} & 0.124 & 0.245 & 0.317 \\

Direct3D-S2~\citep{direct3d}
& \xmark
& 0.074 & 0.122 & 0.247 & 0.314 \\

Hi3DGen~\citep{hi3dgen}
& \xmark
& 0.065 & 0.113 & 0.252 & 0.301 \\

\midrule

CraftsMan 1.5~\citep{craftsman3d}
& \cmark
& 0.074 & 0.129 & 0.237 & 0.298 \\

UniLat3D~\citep{unilat3d}
& \cmark
& 0.071 & 0.119 & 0.252 & 0.307 \\

Hunyuan3D 2.1~\citep{hunyuan3d2.1}
& \cmark
& 0.075 & 0.125 & 0.250 & 0.320 \\

\midrule
\rowcolor[gray]{0.93}
\textbf{Position Forcing (Ours)}
& \cmark
& \textbf{0.077}
& \textbf{0.130}
& \textbf{0.257}
& \textbf{0.321} \\

\bottomrule
\end{tabular}}
\end{table}
\begin{figure}
    \centering
    \includegraphics[width=0.85\linewidth]{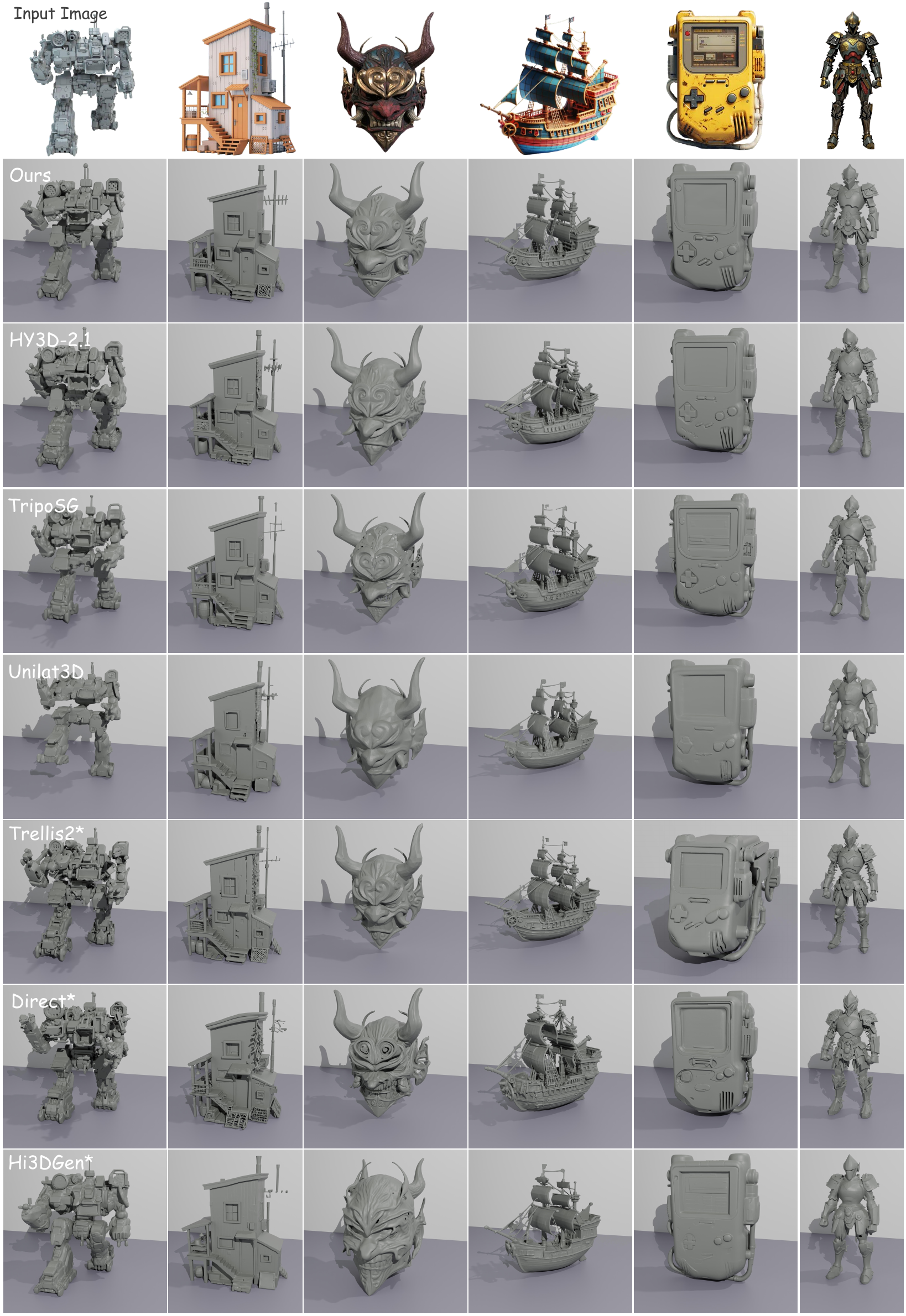}
    \caption{
    Qualitative comparison of image-conditioned 3D generation.
    The input images are shown in the top row.
    Compared with existing methods, our approach produces more faithful 3D geometry with coherent global structures and fine-grained details across diverse object categories. Methods marked with * adopt a two-stage generation framework.
    }
    \label{fig:qualitative}
\end{figure}
We evaluate image-conditioned 3D generation using ULIP~\citep{ulip} and Uni3D~\citep{uni3d} similarities.
We compare against representative existing 3D generation methods, including multi-stage approaches TRELLIS~\citep{trellis}, TRELLIS~2~\citep{trellis2}, Direct3D-S2~\citep{direct3d}, and Hi3DGen~\citep{hi3dgen}, as well as single-stage approaches CraftsMan~1.5~\citep{craftsman3d}, UniLat3D~\citep{unilat3d}, and Hunyuan3D~2.1~\citep{hunyuan3d2.1}.
As shown in Table~\ref{tab:generation}, Position Forcing achieves the best or tied-best performance across all four metrics, outperforming both single-stage and multi-stage baselines.
Qualitative results in Fig.~\ref{fig:qualitative} further demonstrate improved structural consistency and finer geometric details.

\subsection{Ablations and Analysis}
\label{sec:ablation}

\begin{figure*}[t]
    \centering
    \setlength{\abovecaptionskip}{2pt}
    \includegraphics[width=\textwidth]{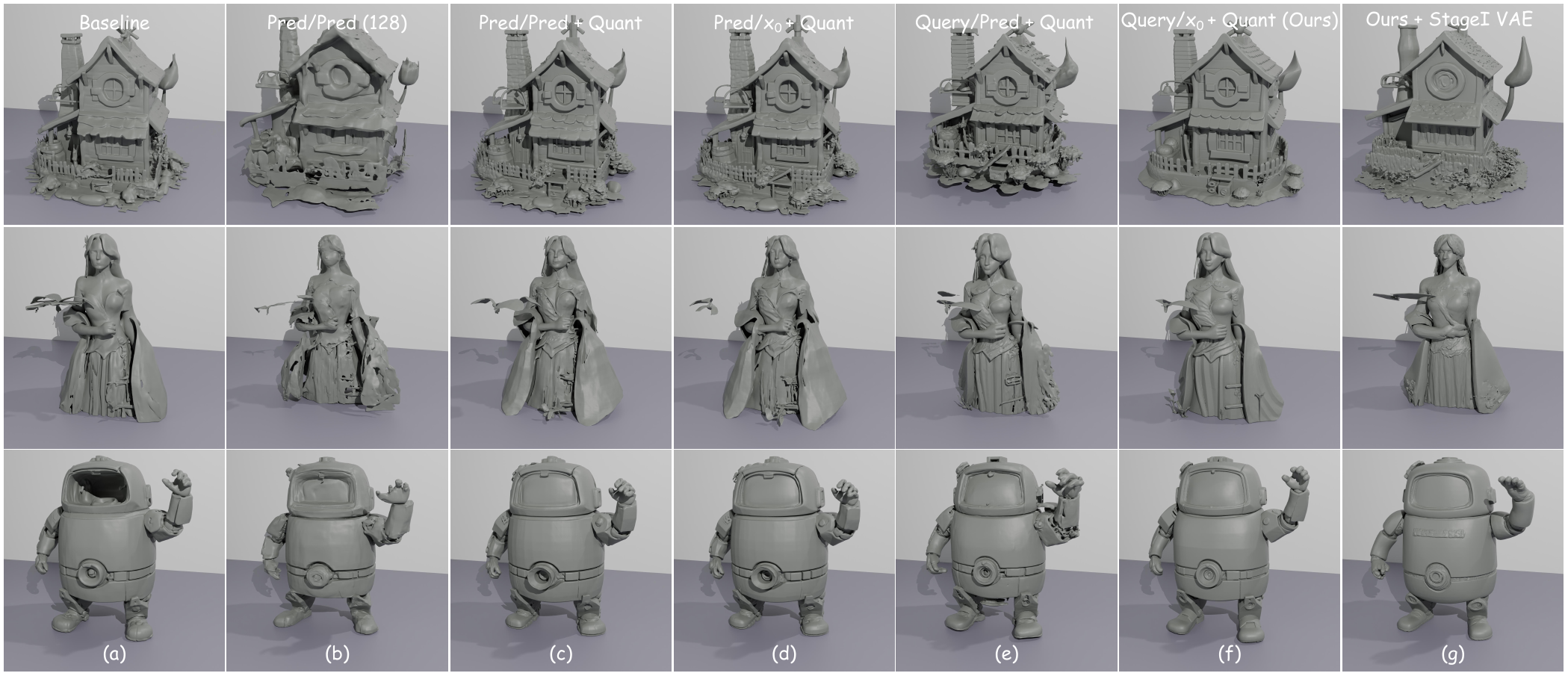}
    \vspace{5pt}
    \begingroup
    \scriptsize
    \setlength{\tabcolsep}{3.8pt}
    \renewcommand{\arraystretch}{1.10}

    \begin{tabular*}{\textwidth}{
        @{\extracolsep{\fill}}
        l
        c c c c c c c
        @{}
    }
        \toprule
        \textbf{Configuration}
        & (a)
        & (b)
        & (c)
        & (d)
        & (e)
        & \textbf{(f) Ours}
        & (g)
        \\
        \midrule

        Train pos.
        & --
        & Pred
        & Pred
        & Pred
        & Query
        & Query
        & Query
        \\

        Infer pos.
        & --
        & Pred
        & Pred
        & $x_0$
        & Pred
        & $x_0$
        & $x_0$
        \\

        Progressive quant.
        & $\times$
        & $\times$
        & $\checkmark$
        & $\checkmark$
        & $\checkmark$
        & $\checkmark$
        & $\checkmark$
        \\

        Resolution
        & --
        & 128
        & Prog.
        & Prog.
        & Prog.
        & Prog.
        & Prog.
        \\

        VAE
        & VetSet
        & Joint
        & Joint
        & Joint
        & Joint
        & Joint
        & Stage-I
        \\

        \midrule

        ULIP-T $\uparrow$
        & 0.075
        & \underline{0.076}
        & 0.074
        & 0.075
        & 0.058
        & \textbf{0.077}
        & \underline{0.076}
        \\

        ULIP-I $\uparrow$
        & 0.125
        & 0.127
        & 0.124
        & 0.126
        & 0.095
        & \textbf{0.130}
        & \underline{0.128}
        \\

        Uni3D-T $\uparrow$
        & 0.250
        & 0.247
        & \underline{0.256}
        & \textbf{0.257}
        & 0.225
        & \textbf{0.257}
        & 0.248
        \\

        Uni3D-I $\uparrow$
        & 0.320
        & 0.305
        & \underline{0.321}
        & \textbf{0.323}
        & 0.265
        & \underline{0.321}
        & 0.308
        \\

        \bottomrule
    \end{tabular*}
    \endgroup
    \vspace{2pt}
    \caption{
        \textbf{Qualitative and quantitative ablation of Position Forcing.}
        Columns (a)--(g) share the same configurations across qualitative
        and quantitative results.
        Query, Pred, and $x_0$ denote ground-truth query positions,
        predicted positions, and positions recovered from the predicted
        clean latent, respectively.
        Prog. denotes progressive position quantization.
        Configuration (g) uses the Stage-I VAE instead of the jointly trained VAE.
        \textbf{Bold} and \underline{underlining} denote the best and
        second-best results.
    }
    \vspace{-12pt}
    \label{fig:ablation}
\end{figure*}

We ablate progressive position quantization, inference-time
position recovery, training-time positional conditioning,
and joint VAE training in Fig.~\ref{fig:ablation}.
Configuration (a) is the VetSet baseline without explicit
positional conditioning, while (f) is our full model.
For each input, all configurations use the same random seed
and sampling settings.

\textbf{Progressive Position Quantization.}
We evaluate progressive position quantization by comparing
(b) and (c).
Configuration (b) uses a fixed position resolution of 128 throughout denoising, whereas (c) progressively
increases the conditioning resolution.
Qualitatively, (b) exhibits fragmented structures around the
house and irregular character surfaces, while (c) reduces
these artifacts and produces better geometry.
These results support using coarse positional conditions
to mitigate unreliable position predictions early in denoising,
followed by progressively finer spatial guidance.

\textbf{Position Recovery from Predicted Clean Latents.}
We first examine the source of position recovery during
inference by comparing (c) and (d).
Both configurations share the same DiT checkpoint and differ
only in the input to position recovery: (c) uses the current
noisy latent, while (d) uses the predicted clean latent $x_0$.
Although (d) improves all four metrics, the gains are modest
because imprecise training-time positions limit the model's
ability to benefit from more accurate positions at inference.
This observation motivates the use of ground-truth query positions
during training to provide a more accurate spatial reference.

\textbf{Ground-Truth Query Conditioning during Training.}
We evaluate ground-truth query conditioning during training
by comparing (d) and (f).
Both configurations use progressive quantization and recover
positions from $x_0$ during inference, while (d) uses predicted
positions during training and (f) uses ground-truth query positions.
Compared with (d), (f) produces cleaner local structures
on the house and more complete character geometry,
while improving both ULIP metrics.
We further compare (e) and (f), which share the same
query-conditioned DiT checkpoint and differ only in
the source of position recovery during inference.
Direct position prediction from noisy latents in (e) still
produces substantial geometric artifacts, whereas recovery
from $x_0$ in (f) markedly improves geometric completeness
and all four metrics.
This comparison suggests that recovering positions from
predicted clean latents provides more accurate positional
estimates, enabling better generation performance for
the model trained with ground-truth queries.

\textbf{Joint VAE Training.}
Finally, we evaluate joint VAE training by comparing
our full configuration (f) with configuration (g).
Configuration (g) replaces the jointly trained VAE with
the Stage-I VAE and retrains the DiT under otherwise
identical settings.
Compared with (g), (f) reduces fragmented structures and better preserves details,
while achieving higher scores on all four metrics,
demonstrating improved generation quality.
Together with the improved position recovery in
Table~\ref{tab:position_recovery}, these results indicate
that joint geometry and position learning yields a more
effective spatially aware representation, benefiting
downstream generation.

\begin{wrapfigure}{r}{0.55\textwidth}
    \centering
    \vspace{-10pt}
    \includegraphics[width=\linewidth]{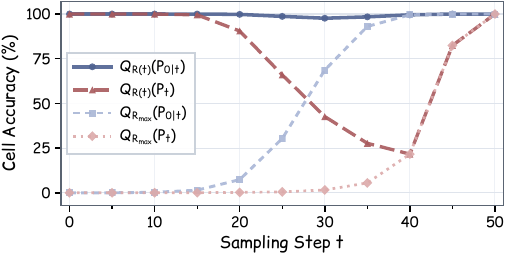}
    \vspace{-16pt}
    \caption{
\textbf{Positional consistency during sampling.}
We compare cell accuracy along the sampling trajectory
for positions recovered from predicted clean latents
($P_{0\mid t}$) and noisy latents ($P_t$).
Accuracy is the percentage of tokens whose intermediate
and final decoded positions fall into the same cell,
using progressive resolution $R(t)$ or fixed resolution
$R_{\max}$ for both.
    }
    \vspace{-5pt}
    \label{fig:position_consistency}
\end{wrapfigure}

\textbf{Analysis of Positional Feedback.}
We analyze the consistency between recovered and final
generated positions during sampling to understand the roles
of progressive quantization and clean-latent position recovery.
As shown in Fig.~\ref{fig:position_consistency}, progressive
quantization improves overall accuracy across the sampling
trajectory.
This enables coarse spatial conditioning before fine-grained
positions become reliable, consistent with the improved
structural integrity in Fig.~\ref{fig:ablation}(b)--(c).
Compared with $P_t$, $P_{0\mid t}$ aligns with the final
positions earlier and maintains high agreement under
progressive quantization, consistent with the substantial
gains in Fig.~\ref{fig:ablation}(e)--(f).
Together, these designs provide coarse-to-fine spatial
guidance that better aligns with the final generated positions.


\section{Conclusions}

We presented Position Forcing, a position-based self-conditioning
framework for single-stage 3D generation.
Our key observation is that VecSet latents retain
correspondences with their spatial queries, enabling positional guidance
to be recovered directly from geometry latents.
Position VAE enables position recovery from geometry latents, while
Position-Forced DiT uses the recovered positions as spatial guidance
during denoising.
Experiments demonstrate competitive performance against both
single-stage and multi-stage methods.

\bibliography{iclr2027_conference}
\bibliographystyle{iclr2027_conference}

\clearpage
\appendix



\newpage
\appendix

\section{Appendix}
\subsection{Implementation Details}
\label{sec:implementation}

\paragraph{Position VAE.}
Position VAE uses latent tokens with 64 channels.
The geometry encoder and decoder contain 8 and 16 Transformer
layers, respectively, while the position decoder contains
4 Transformer layers.
During VAE training, we perturb the query coordinates before
positional encoding, independently sampling each offset
component from $\mathcal{U}(-1/64,1/64)$.
We first train the position decoder with the pretrained VAE
frozen, then jointly finetune the encoder, geometry decoder,
and position decoder.
The position loss and KL regularization weights are set to
$\lambda_{\mathrm{pos}}=1$ and $\lambda_{\mathrm{KL}}=10^{-6}$,
respectively.

\paragraph{Position-Forced DiT.}
We use a FLUX-based Transformer architecture~\citep{flux} with 12 double-stream blocks
and 24 single-stream blocks, a hidden dimension of 1536,
16 attention heads, and an MLP expansion ratio of 4.
For 3D RoPE, we allocate 32 channels to each spatial axis
and set the frequency base to 10,000.
Image conditions are extracted using DINOv2-Giant, concatenated with the geometry tokens, and jointly processed by the FLUX-based Transformer.
Training images have a resolution of $518\times518$,
and image conditions are dropped with probability 0.1
to support classifier-free guidance.

\paragraph{DiT Training.}
Position VAE remains frozen throughout DiT training.
We first train the DiT with 4096 latent tokens and
subsequently continue training with 6144 tokens.
Timesteps for flow-matching training are sampled uniformly
from $[0,1]$.
We use AdamW with a base learning rate of $2\times10^{-5}$,
$\beta_1=0.9$, $\beta_2=0.99$, weight decay of 0.01,
and $\epsilon=10^{-6}$.
The learning rate follows a 500-step warmup followed by
cosine decay, and gradients are clipped to a maximum
norm of 1.0.

\paragraph{Position Quantization and Perturbation.}
For the progressive resolution schedule in
Eq.~\ref{eq:resolution_schedule}, we set
$m_{\min}=-1$, $m_{\max}=9$, $R_{\min}=1$, and $R_{\max}=128$.
This schedule maintains a single spatial cell at high noise
levels, progressively refines the grid, and retains the
maximum resolution during the final part of denoising.
Quantized coordinates at all resolutions are mapped to
a common reference coordinate system with resolution 128.
During DiT training, positional perturbation is activated
independently for each sample with probability 0.5
and applied after quantization and coordinate mapping.
Each coordinate offset takes a value in $\{-1,0,1\}$,
with a fixed magnitude in this reference coordinate system
regardless of the current quantization resolution.
Perturbations are disabled when $R=1$ so that all tokens
share the same positional encoding, and are not used
during inference.

\paragraph{Inference.}
At inference time, we use 12288 latent tokens and
input images of resolution $1022\times1022$.
We perform 50-step Euler sampling with classifier-free
guidance (CFG) at a guidance scale of 5.0, using BF16 precision.
The initial step uses $R=1$, and subsequent steps obtain
positional conditions from the clean latent prediction
of the preceding step.
The conditional and unconditional CFG branches share
the same positional conditions.
The generated latents are decoded by the geometry decoder
of Position VAE.
Meshes are extracted using the \texttt{dmc} mode
at a resolution of 512 with an isosurface threshold of 0.

\subsection{Training and Inference Algorithms}
\label{sec:algorithms}

Algorithm~\ref{alg:position_forcing} summarizes the core
training and inference procedures of Position Forcing.
During training, we quantize the query positions associated
with the frozen VAE encoder to provide spatial guidance
to the DiT.
During inference, positions recovered from the predicted
clean latents are quantized at the next step's resolution
to condition subsequent denoising.
Here, \texttt{rope\_idx} denotes the quantized coordinates
used to construct 3D RoPE within the DiT.
For clarity, the pseudocode omits training-time positional
perturbations and image-condition dropout.

\begin{algorithm}[t]
\caption{Position Forcing training and inference in a
PyTorch-like style.}
\label{alg:position_forcing}
\begin{lstlisting}[style=pytorch]
# vae.encode: returns latents and their query positions
# quantize: Q_R, including quantization and coordinate mapping
# dit: constructs 3D RoPE internally from rope_idx
# dit_cfg: applies image CFG with shared position conditions

# Training: use query positions from the frozen encoder
def training_loss(point_cloud, image):
    with no_grad():
        z0, query_pos = vae.encode(point_cloud)
        image_cond = image_encoder(image)

    t, noise = sample_time_and_noise(z0)
    zt = (1 - t) * z0 + t * noise

    # Quantize query positions at the current resolution
    pos = quantize(query_pos, R(t))
    v = dit(zt, t, image_cond, rope_idx=pos)

    return mse(v, noise - z0)


# Inference: recover positions from predicted clean latents
@no_grad()
def sample(noise, image, times):
    z = noise
    image_cond = image_encoder(image)

    # Initially R = 1: all tokens share the same coordinates
    pos = zeros_like_positions(z)

    for t, t_next in consecutive_pairs(times):
        # Predict velocity using current position conditions
        v = dit_cfg(z, t, image_cond, rope_idx=pos)

        # Predict clean latents and recover token positions
        z0_pred = z - t * v
        predicted_pos = pos_decoder(z0_pred)

        # Prepare position conditions for the next step
        pos = quantize(predicted_pos, R(t_next))

        # Advance sampling with an Euler update
        z = z + (t_next - t) * v

    # Decode the generated geometry
    return mesh_decoder(z)
\end{lstlisting}
\end{algorithm}

\subsection{Additional Qualitative Results}
\label{sec:additional_results}

\begin{figure}[p]
    \centering
    \includegraphics[width=0.85\linewidth]{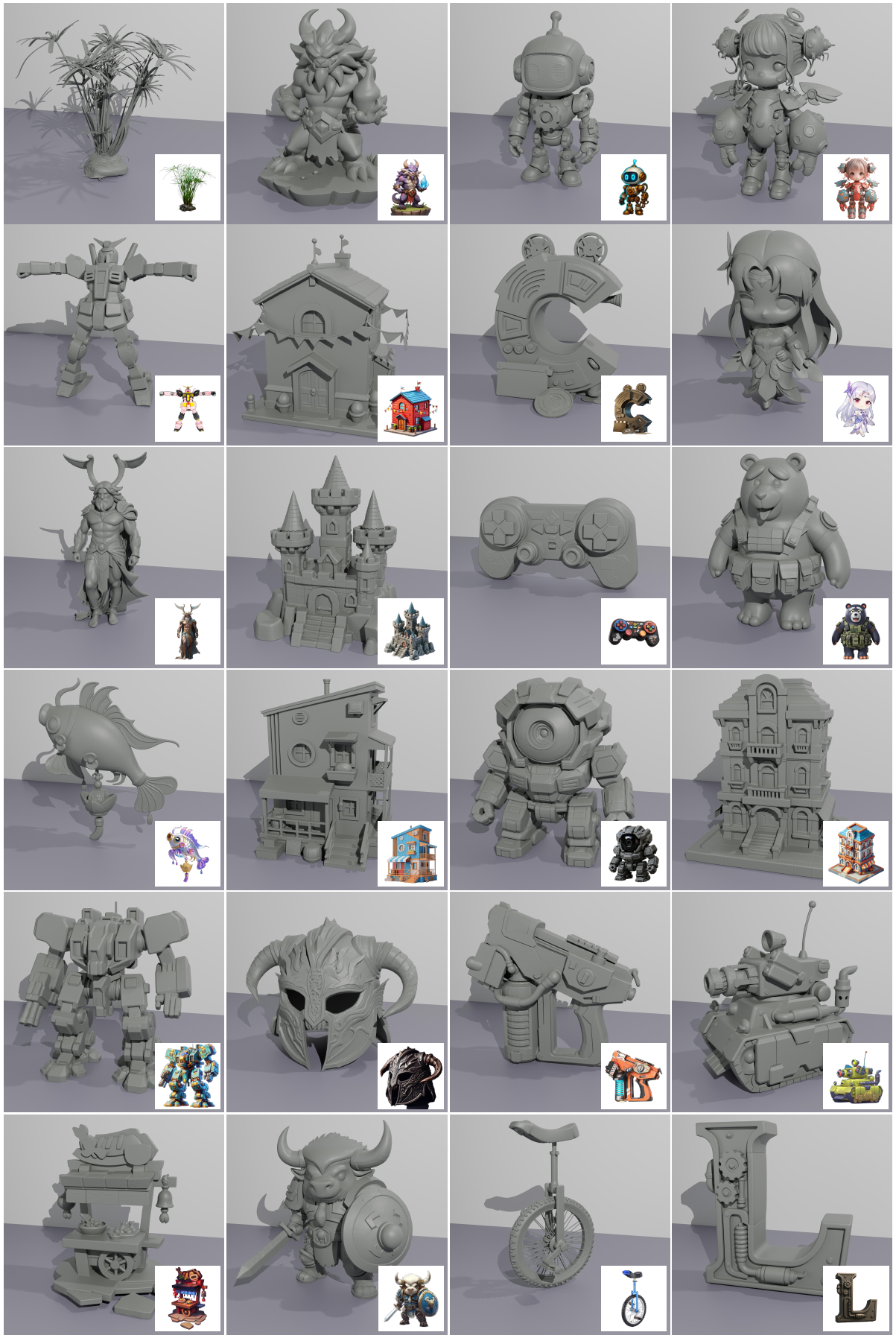}
    \caption{
        \textbf{Additional generation results.}
        Generated geometries are shown with their corresponding
        input images inset at the bottom right.
    }
    \label{fig:more_results}
\end{figure}

Fig.~\ref{fig:more_results} presents additional image-conditioned
3D generation results across diverse categories, including
plants, stylized characters, buildings, and mechanical objects.
These examples illustrate the ability of Position Forcing
to capture both overall shape and local geometric details,
such as thin plant leaves, character ornaments,
architectural elements, and mechanical components.

\subsection{Multi-view Qualitative Comparison}
\label{sec:multiview_comparison}

\begin{figure}[t]
    \centering
    \includegraphics[width=0.9\linewidth]{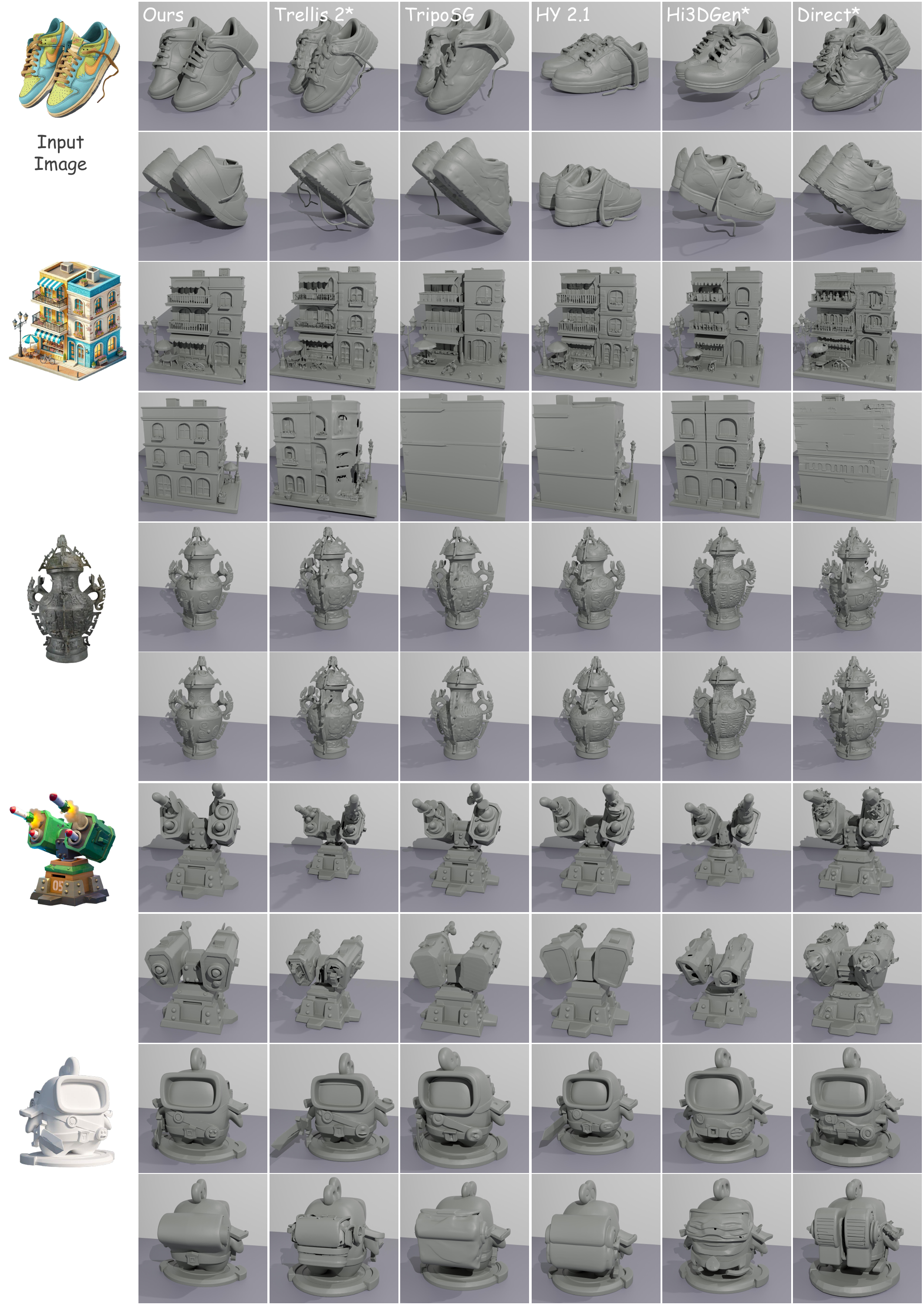}
    \caption{
    Multi-view qualitative comparison of image-conditioned 3D generation.
    Position Forcing produces smoother surfaces and more geometrically
    plausible details, particularly in rear views, with fewer
    surface artifacts and structural distortions. Methods marked with * adopt a two-stage generation framework.
    }
    \label{fig:multiview}
\end{figure}

Fig.~\ref{fig:multiview} presents generated 3D shapes from multiple
viewpoints to further assess the geometric quality of regions
not visible in the conditioning images.
Compared with the baseline, Position Forcing produces smoother
surfaces and more geometrically plausible local details in rear
views, including more regular building windows, better-preserved
mechanical components, and cleaner character backs.
These results further demonstrate the advantages of Position Forcing
in surface quality and local geometric detail.

\subsection{Generation Trajectory Visualization}
\label{sec:generation_trajectory}

\begin{figure}[h]
    \centering
    \includegraphics[width=\textwidth]{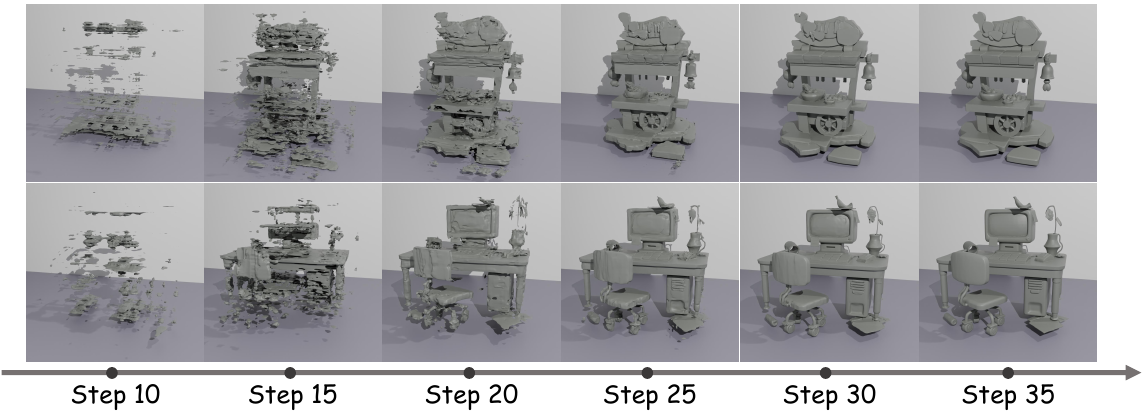}
    \caption{
    \textbf{Generation trajectory visualization.}
    We use a fixed 50-step sampling schedule and visualize the predicted
    clean sample $\hat{x}_0$ at different generation steps.
    The intermediate predictions show that the model progressively forms
    the global structure in the early steps and then refines local details
    in later steps, eventually converging to the final output.
    }
    \label{fig:generation_trajectory}
\end{figure}

We further visualize the generation trajectory of our model in
Fig.~\ref{fig:generation_trajectory}.
We use a fixed 50-step sampling schedule and show the predicted clean
sample $\hat{x}_0$ at selected generation steps.
As the generation proceeds, the model first establishes the global
structure and rough appearance of the output, and then gradually refines
local details, following a coarse-to-fine generation process.


\end{document}